\documentclass[10pt,twocolumn,letterpaper]{article}

\usepackage{iccv}
\usepackage{times}
\usepackage{epsfig}
\usepackage{graphicx}
\usepackage{amsmath}
\usepackage{amssymb}
\usepackage{multicol}
\usepackage{multirow}
\usepackage{color, colortbl}

\graphicspath{{./images}}

\usepackage[dvipsnames]{xcolor}

\usepackage[pagebackref=true,breaklinks=true,letterpaper=true,colorlinks,bookmarks=false]{hyperref}

\def\iccvPaperID{10474} 
\def\confYear{ICCV 2021}
\iccvfinalcopy

\newcommand\ntxt[1]{\textcolor{black}{#1}}
\newcommand\ntext[1]{\textcolor{black}{#1}}
\definecolor{Gray}{gray}{0.9}
\definecolor{Gray2}{gray}{0.8}
\definecolor{Gray3}{gray}{0.7}
\definecolor{Gray4}{gray}{0.6}

\newcommand{\sidetext}[1]{\begin{tabular}{@{}c@{}}#1\end{tabular}}

\begin{document}
\title{DR{\AE}M -- A discriminatively trained reconstruction embedding for surface anomaly detection}

\author{Vitjan Zavrtanik \and Matej Kristan \and Danijel Skočaj \and
University of Ljubljana, Faculty of Computer and Information Science \\
{\tt\small \{vitjan.zavrtanik, matej.kristan, danijel.skocaj\}@fri.uni-lj.si}
}

%

\maketitle

\begin{abstract}

Visual surface anomaly detection aims to detect local image regions that significantly deviate from normal appearance. Recent surface anomaly detection methods rely on generative models to accurately reconstruct the normal areas and to fail on anomalies. These methods are trained only on anomaly-free images, and often require hand-crafted post-processing steps to localize the anomalies, which prohibits optimizing the feature extraction for maximal detection capability. In addition to reconstructive approach, we cast surface anomaly detection primarily as a discriminative problem and propose a discriminatively trained reconstruction anomaly embedding model (DR{\AE}M). The proposed method learns a joint representation of an anomalous image and its anomaly-free reconstruction, while simultaneously learning a decision boundary between normal and anomalous examples. The method enables direct anomaly localization without the need for additional complicated post-processing of the network output and can be trained using simple and general anomaly simulations. On the challenging MVTec anomaly detection dataset, DR{\AE}M outperforms the current state-of-the-art unsupervised methods by a large margin and even delivers detection performance close to the fully-supervised methods on the widely used DAGM surface-defect detection dataset, while substantially outperforming them in localization accuracy. 




\end{abstract}

\section{Introduction}

\begin{figure}
\centering
  \includegraphics[width=1.0\linewidth]{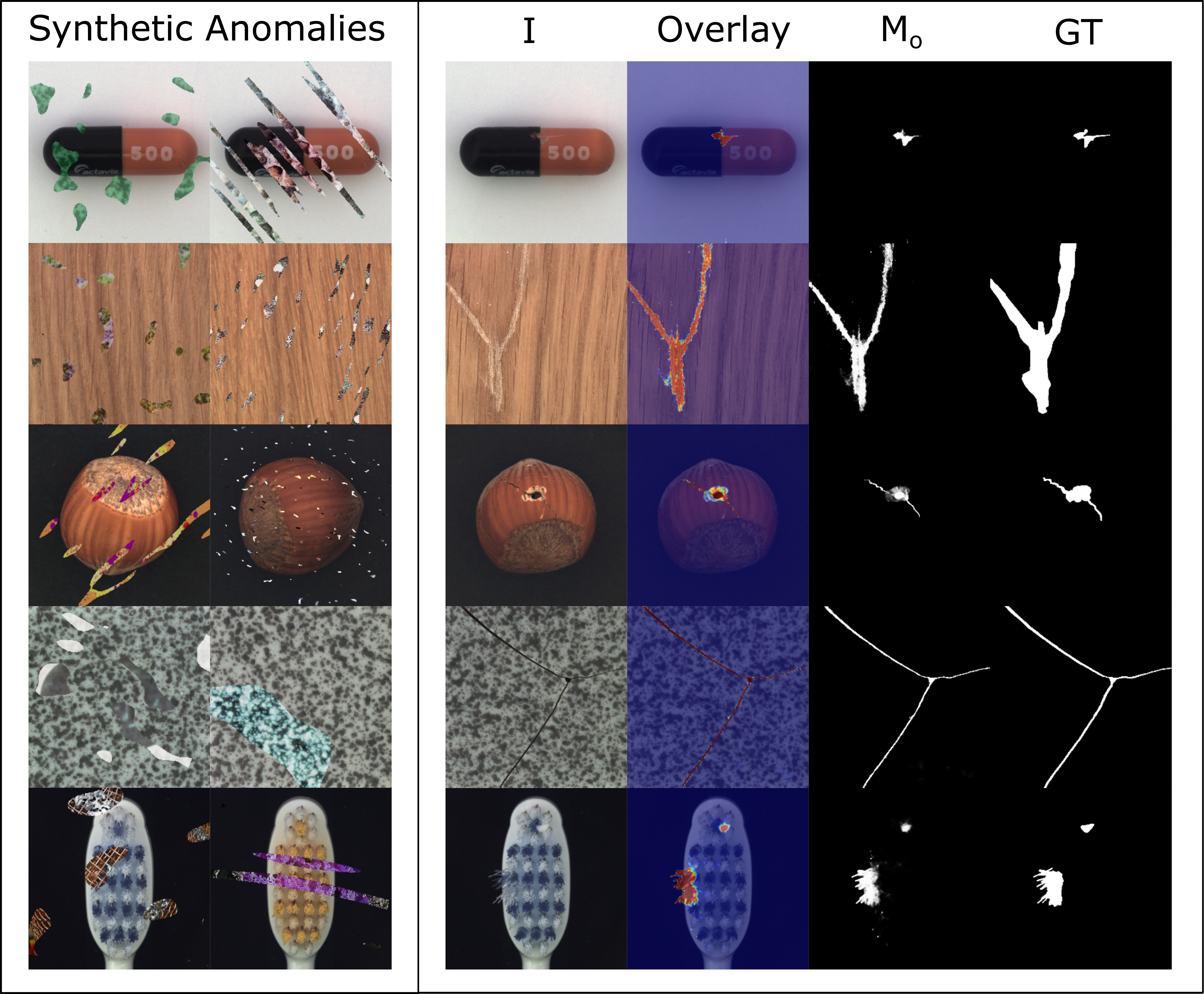}
\caption{DR{\AE}M estimates the decision boundary between the normal an anomalous pixels solely by training on synthetic anomalies automatically generated on anomaly-free images (left) and generalizes to a variety of real-world anomalies (right). The result ($M_o$) closely matches the ground truth (GT).
}
\label{fig:anomaly_detection_example}
\end{figure}

\ntext{Surface anomaly detection addresses localization of image regions that deviate from a normal appearance (Figure~\ref{fig:anomaly_detection_example}). 
A closely related general anomaly detection problem considers anomalies as \textit{entire images} that significantly differ from the non-anomalous training set images. In contrast, in surface anomaly detection problems, the anomalies occupy only \textit{a small fraction} of image pixels and are typically close to the training set distribution. This is a particularly challenging task, which is common in quality control and surface defect localization applications.}

\ntext{In practice, anomaly appearances may significantly vary, and in applications like quality control, images with anomalies present are rare and manual annotation may be overly time consuming. This leads to highly imbalanced training sets, often containing only anomaly-free images. Significant effort has thus been recently invested in designing robust surface anomaly detection methods that preferably require minimal supervision from manual annotation.}

\begin{figure}
\centering
  \includegraphics[width=1.0\linewidth]{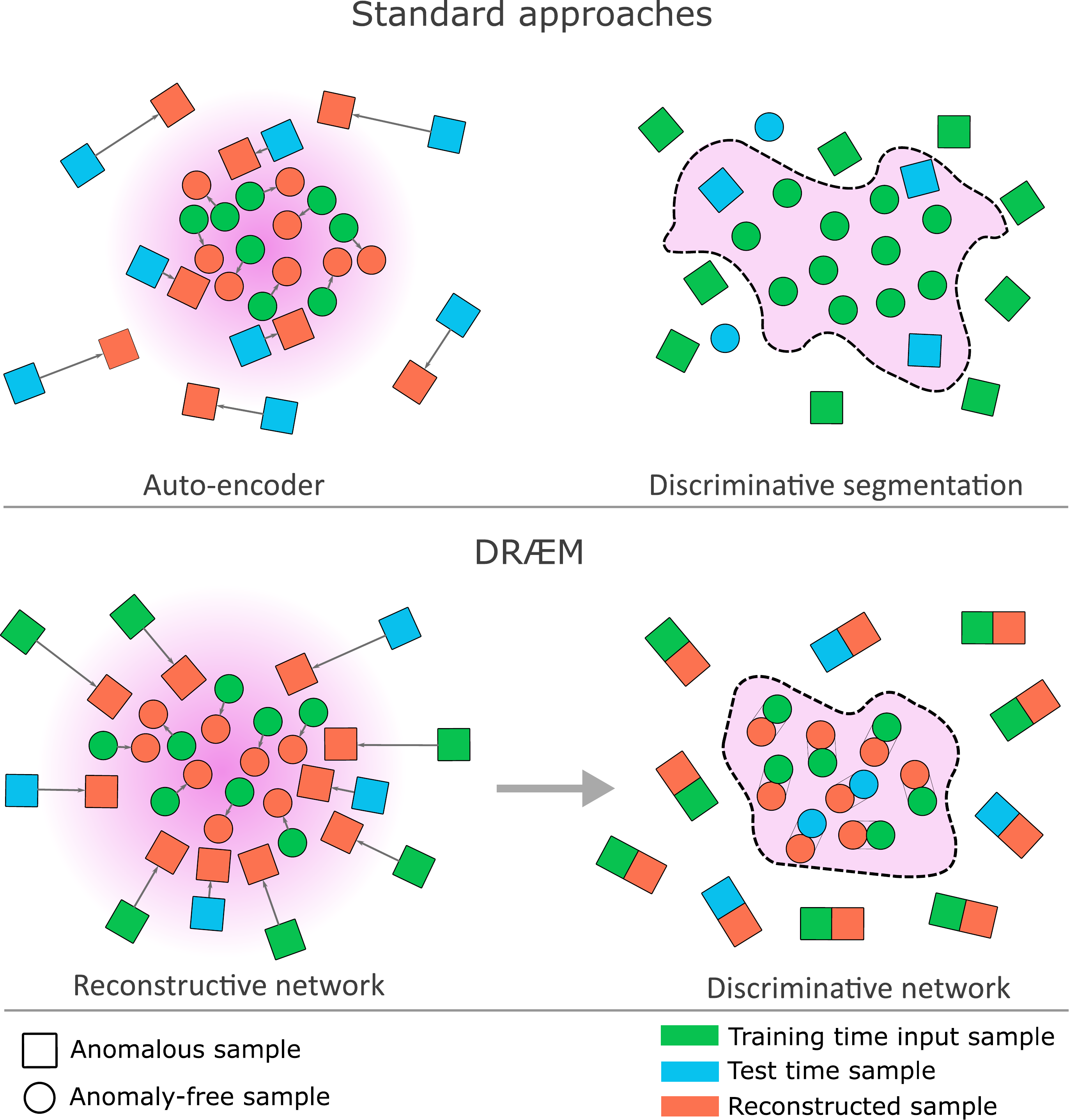}
\caption{Autoencoders over-generalize to anomalies, while discriminative approaches over-fit to the synthetic anomalies and do not generalize to real data. Our approach jointly discriminatively learns the reconstruction subspace and a hyper-plane over the joint original and reconstructed space using the simulated anomalies and leads to substantially better generalization to real anomalies.
%
}
\label{fig:point_examples}
\end{figure}

Reconstructive methods, such as Autoencoders~\cite{bergmann2018improving,akcay2018ganomaly,akccay2019skip, tang2020anomaly} and GANs~\cite{schlegl2017unsupervised,schlegl2019f}, have been extensively explored since they enable learning of a powerful reconstruction subspace, using only anomaly-free images.
Relying on poor reconstruction capability of anomalous regions, not observed in training, the anomalies can then be detected by thresholding the difference between the input image and its reconstruction. 
However, determining the presence of anomalies that are not substantially different from normal appearance remains challenging, since these are often well reconstructed, as depicted in Figure \ref{fig:point_examples}, top-left.

Recent improvements thus consider the difference between deep features extracted from a general-purpose network and a network specialized for anomaly-free images~\cite{bergmann2020uninformed}. Discrimination can also be formulated as a deviation from a dense clustering of non-anomalous textures within the deep subspace~\cite{pmlr-v80-ruff18a,chalapathy2018anomaly}, as forming such a compact subspace prevents anomalies from being mapped close to anomaly-free samples.
A common drawback of the generative methods is that they only learn the model from anomaly-free data, and are not explicitly optimized for discriminative anomaly detection, since positive examples (i.e., anomalies) are not available at training time. Synthetic anomalies could be considered to train discriminative segmentation methods~\cite{chen2017rethinking,ronneberger2015unet}, but this leads to over-fitting to synthetic appearances and results in a learned decision boundary that generalizes poorly to real anomalies (Figure \ref{fig:point_examples}, top-right).

We hypothesize that over-fitting can be substantially reduced by training a discriminative model over the joint, reconstructed and original, appearance along with the reconstruction subspace. This way the model does not overfit to synthetic appearance, but rather learns a local-appearance-conditioned distance function between the original and reconstructed anomaly appearance, which generalizes well over a range of real anomalies (see Figure~\ref{fig:point_examples}, bottom).

\begin{figure*}
\centering
  \includegraphics[width=0.80\linewidth]{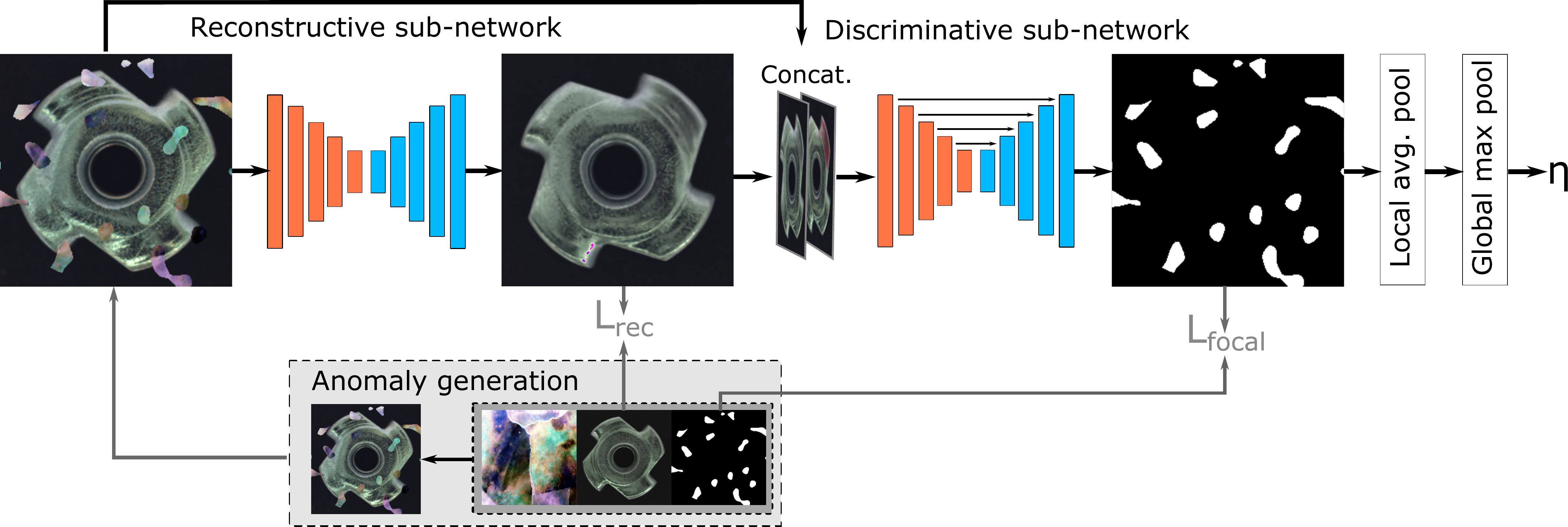}
\caption{The anomaly detection process of the proposed method. First anomalous regions are implicitly detected and inpainted by the reconstructive sub-network trained using $L_{rec}$. The output of the reconstructive sub-network and the input image are then concatenated and fed into the discriminative sub-network. The segmentation network, trained using the Focal loss $L_{focal}$\cite{lin2017focal}, localizes the anomalous region and produces an anomaly map. The image level anomaly score $\eta$ is acquired from the anomaly score map.
}
\label{fig:process}
\end{figure*}

To validate our hypothesis, we propose, as our main contribution, a new deep surface anomaly detection network, discriminatively trained in an end-to-end manner on \ntext{synthetically generated just-out-of-distribution patterns, which do not have to faithfully represent the target-domain anomalies.}
The network is composed of a reconstructive sub-network, followed by a discriminative sub-network (Figure~\ref{fig:process}). The reconstructive sub-network is trained to learn anomaly-free reconstruction, while the discriminative sub-network learns a discriminative model over the joint appearance of the original and reconstructed images, producing a high-fidelity per-pixel anomaly detection map (Figure~\ref{fig:anomaly_detection_example}). 

In contrast to related approaches that learn surrogate generative tasks, the proposed model is trained discriminatively
, yet does not require the synthetic anomaly appearances to closely match the anomalies at test time and outperforms the recent, more complex, state-of-the-art methods by a large margin.

\section{Related work}
Many surface anomaly detection methods focus on image reconstruction and detect anomalies based on image reconstruction error \cite{akcay2018ganomaly,akccay2019skip,bergmann2018improving,schlegl2017unsupervised,schlegl2019f,tang2020anomaly,zavrtanik2020riad}. Auto-encoders are commonly used for image reconstruction~\cite{bergmann2018improving}. In \cite{akcay2018ganomaly,akccay2019skip,tang2020anomaly} auto-encoders are trained with adversarial losses. The anomaly score of the image is then based on the image reconstruction quality or in the case of adversarially trained auto-encoders, the discriminator output. In \cite{schlegl2017unsupervised,schlegl2019f} a GAN~\cite{goodfellow2014generative} is trained to generate images that fit the training distribution. In \cite{schlegl2019f} an encoder network is additionally trained that finds the latent representation of the input image that minimizes the reconstruction loss when used as the input by the pretrained generator. The anomaly score is then based on the reconstruction quality and the discriminator output. In \cite{wu2020mirrored} an interpolation auto-encoder is trained to learn a dense representation space of in-distribution samples. The anomaly score is then based on a discriminator, trained to estimate the distance between the input-input and input-output joint distributions, however the approach to surface anomaly detection remains generative as the discriminator evaluates the reconstruction quality.

\ntxt{Instead of the commonly used image space reconstruction, the reconstruction of pretrained network features can also be used for surface anomaly detection~\cite{bergmann2020uninformed,dfrAD}. Anomalies are detected based on the assumption that features of a pre-trained network will not be faithfully reconstructued by another network trained only on anomaly-free images.  Alternatively~\cite{madGauss,defard2020padim} propose surface anomaly detection as identifying significant deviations from a Gaussian fitted to anomaly-free features of a pre-trained network. This requires a unimodal distribution of the anomaly-free visual features which is problematic on diverse datasets.} \cite{liu2020towards} propose a one-class variational auto-encoder gradient-based attention maps as output anomaly maps. However the method is sensitive to subtle anomalies close to the normal sample distribution. 


\ntxt{Recently Patch-based one-class classification methods have been considered for surface anomaly detection~\cite{yi2020patch}. These are based on one-class methods~\cite{pmlr-v80-ruff18a, chalapathy2018anomaly} which attempt to estimate a decision boundary around anomaly-free data that separates it from anomalous samples by assuming a unimodal distribution of the anomaly-free data. This assumption is often violated in surface anomaly data.}


\section{DR{\AE}M}

The proposed discriminative joint reconstruction-anomaly embedding method (DR{\AE}M) is composed from a reconstructive and a discriminative sub-networks (see Figure~\ref{fig:process}). The reconstructive sub-network is trained to implicitly detect and reconstruct the anomalies with semantically plausible anomaly-free content, while keeping the non-anomalous regions of the input image unchanged.
Simultaneously, the discriminative sub-network learns a joint reconstruction-anomaly embedding and produces accurate anomaly segmentation maps from the concatenated reconstructed and original appearance. Anomalous training examples are created by a conceptually simple process that simulates anomalies on anomaly-free images. This anomaly generation method provides an arbitrary amount of anomalous samples as well as pixel-perfect anomaly segmentation maps which can be used for training the proposed method without real anomalous samples. 

\subsection{Reconstructive sub-network}

The reconstructive sub-network is formulated as an encoder-decoder architecture that converts the local patterns of an input image into patterns closer to the distribution of normal samples. 
\ntext{The network is trained to reconstruct the original image $I$ from an artificially corrupted version $I_a$ obtained by a simulator (see Section~\ref{sec:generation}).} 
 

An $l_2$ loss is often used in reconstruction based anomaly detection methods \cite{akcay2018ganomaly, akccay2019skip}, however this assumes an independence between neighboring pixels, therefore a patch based SSIM \cite{wang2004image} loss is additionally used as in \cite{bergmann2018improving, zavrtanik2020riad}:
\begin{equation}
L_{SSIM}(I, I_r) = \frac{1}{N_p} \sum_{i=1}^H \sum_{j=1}^W 1-SSIM\big(I,I_r\big)_{(i,j)},
\end{equation}
where $H$ and $W$ are the height and width of image $I$, respectively. $N_p$ is equal to the number of pixels in $I$. $I_r$ is the reconstructed image output by the network. $SSIM(I, I_r)_{(i,j)}$ is the SSIM value for patches of $I$ and $I_r$, centered at image coordinates $(i,j)$. The reconstruction loss is therefore:
\begin{equation}
    L_{rec}(I, I_r) = \lambda L_{SSIM}(I,I_r) + l_{2}(I, I_r),
\label{eq:lrec}
\end{equation}
where $\lambda$ is a loss balancing hyper-parameter.

Note that an additional training signal is acquired from the downstream discriminative network (Section~\ref{sec:discriminative}), which performs anomaly localization by detecting the reconstruction difference. 


\subsection{Discriminative sub-network}\label{sec:discriminative}

The discriminative sub-network uses U-Net~\cite{ronneberger2015unet}-like architecture. The sub-network input $I_c$ is defined as the channel-wise concatenation of the reconstructive sub-network output $I_r$ and the input image $I$. Due to the normality-restoring property of the reconstructive sub-network, the joint appearance of $I$ and $I_r$ differs significantly in anomalous images, providing the information necessary for anomaly segmentation. In reconstruction-based anomaly detection methods anomaly maps are obtained using similarity functions such as SSIM~\cite{wang2004image} to compare the original image to its reconstruction, however a surface anomaly detection-specific similarity measure is difficult to hand-craft. In contrast, the discriminative sub-network learns the appropriate distance measure automatically. The network outputs an anomaly score map $M_o$ of the same size as $I$. Focal Loss \cite{lin2017focal} ($L_{seg}$) is applied on the discriminative sub-network output to increase robustness towards accurate segmentation of hard examples. 

Considering both the segmentation and the reconstructive objectives of the two sub-networks, the total loss used in training DR{\AE}M is
\begin{equation}
    L(I,I_r, M_a, M) = L_{rec}(I,I_r) + L_{seg}(M_a, M),
\end{equation}
where $M_a$ and $M$ are the ground truth and the output anomaly segmentation masks, respectively.


\subsection{Simulated anomaly generation}\label{sec:generation}

\ntext{DR{\AE}M does not require simulations to realistically reflect the real anomaly appearance in the target domain, but rather to generate just-out-of-distribution appearances, which allow learning the appropriate \textit{distance function} to recognize the anomaly by its deviation from normality. The proposed anomaly simulator follows this paradigm.}

\begin{figure}[!htb]
\centering
  \includegraphics[width=0.8\linewidth]{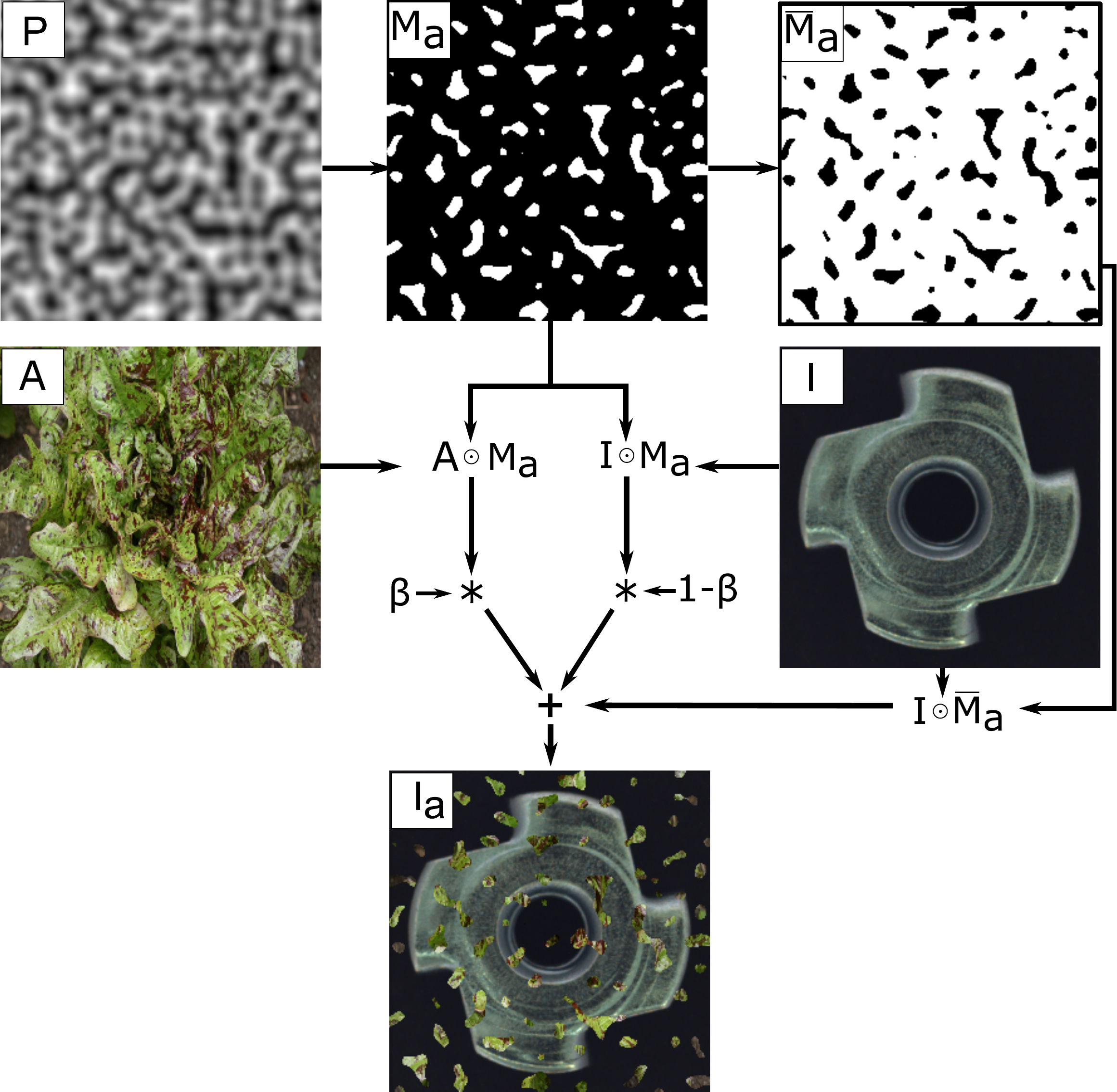}
\caption{Simulated anomaly generation process. The binary anomaly mask $M_a$ is generated from Perlin noise $P$. The anomalous regions are sampled from $A$ according to $M_a$ and placed on the anomaly free image $I$ to generate the anomalous image $I_a$.}
\label{fig:aug}
\end{figure}

\begin{figure}
\centering
  \includegraphics[width=0.9\linewidth]{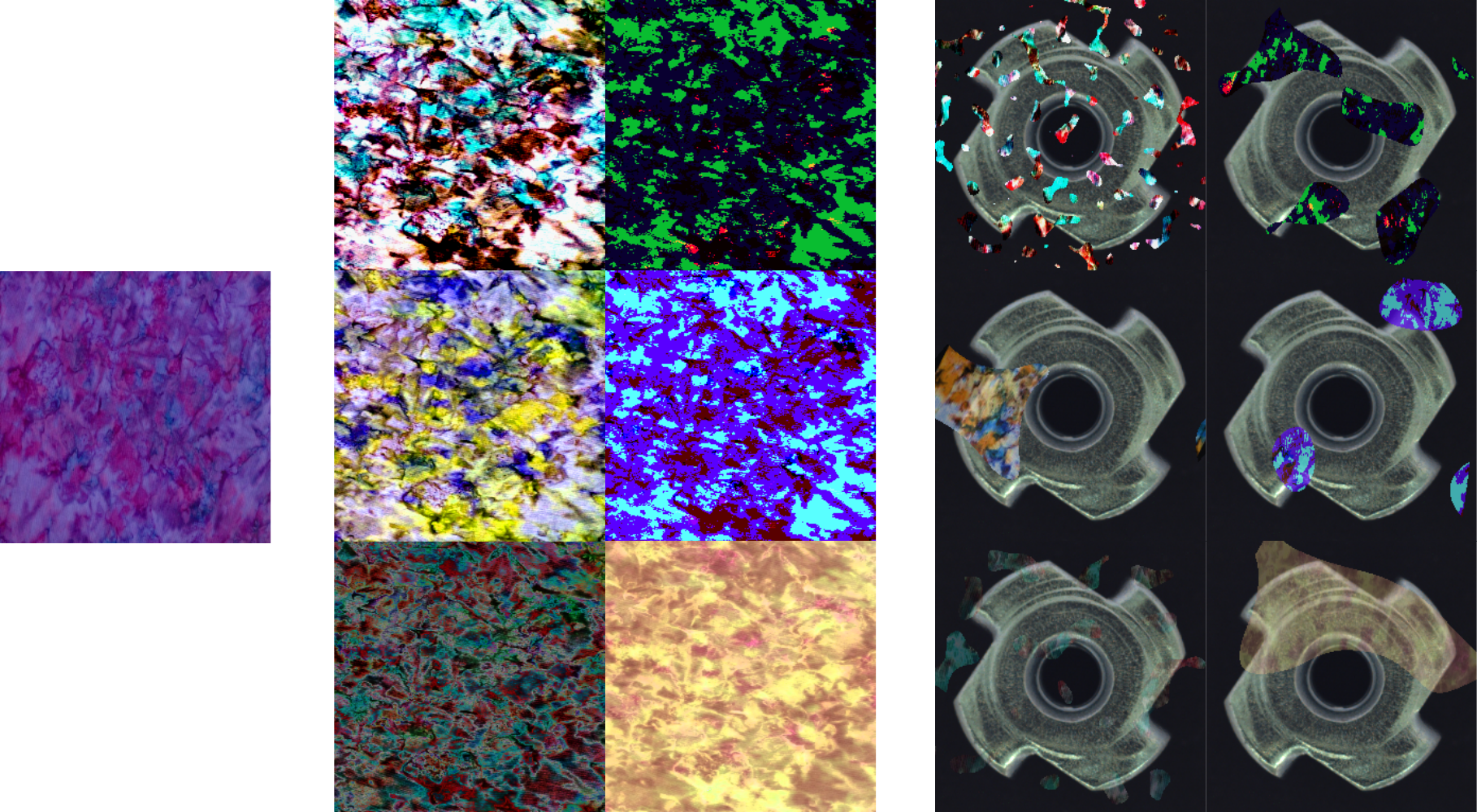}
\caption{The original anomaly source image (left) can be augmented several times (center) to generate a wide variety of simulated anomalous regions (right).}
\label{fig:aug_examples}
\end{figure}

\ntext{A noise image is generated by a Perlin noise generator~\cite{perlin1985image} to capture a variety of anomaly shapes (Figure \ref{fig:aug}, $P$) and binarized by a threshod sampled uniformly at random (Figure \ref{fig:aug}, $M_a$) into an anomaly map $M_a$.}
\ntext{The anomaly texture source image $A$ is sampled from an anomaly source image dataset which is unrelated to the input image distribution (Figure \ref{fig:aug}, $A$).
Random augmentation sampling, inspired by RandAugment~\cite{cubuk2019randaugment}, is then applied by a set of $3$ random augmentation functions sampled from the set: $\{$\textit{posterize, sharpness, solarize, equalize, brightness change, color change, auto-contrast}$\}$.}
\ntext{The augmented texture image $A$ is masked with the anomaly map $M_a$ and blended with $I$ to create anomalies that are just-out-of-distribution, and thus help tighten the decision boundary in the trained network.}
The augmented training image $I_a$ is therefore defined as
\begin{equation}
I_a = \overline{M}_a \odot I + (1-\beta) (M_a \odot I) + \beta (M_a \odot A),
\end{equation}
%
where $\overline{M}_a$ is the inverse of $M_a$, $\odot$ is the element-wise multiplication operation and \ntext{$\beta$ is the opacity parameter in blending. This parameter is sampled uniformly from an interval, i.e., $\beta \in [0.1,1.0]$. The randomized blending and augmentation afford generating diverse anomalous images from as little as \textit{a single texture} (see Figure \ref{fig:aug_examples}).}



 

    
The above described simulator thus generates training sample triplets containing the original anomaly-free image $I$, the augmented image containing simulated anomalies $I_a$ and the pixel-perfect anomaly mask $M_a$. 

\subsection{Surface anomaly localization and detection}
\ntext{The output of the discriminative sub-network is a pixel-level anomaly detection mask $M_o$, which can be interpreted in a straight-forward way for the image-level anomaly score estimation, i.e., whether an anomaly is present in the image.}

First, $M_o$ is smoothed by a mean filter convolution layer to aggregate the local anomaly response information. The final image-level anomaly score $\eta$ is computed by taking the maximum value of the smoothed anomaly score map:
\begin{equation}
    \eta = max\big(M_o \ast  f_{s_f \times s_f}\big)\hspace{0.2cm},
    \label{eq:G}
\end{equation}
where $f_{s_f \times s_f}$ is a mean filter of size $s_f \times s_f$ and $\ast$ is the convolution operator. \ntext{In a preliminary study, we trained a classification network for the image-level anomaly classification, but did not observe improvements over the direct score estimation method (\ref{eq:G}).}


\section{Experiments}
\ntext{DR{\AE}M is extensively evaluated and compared with the recent state-of-the-art on  unsupervised \ntext{surface} anomaly detection and localization. Additionally, individual components of the proposed method and the effectiveness of training on simulated anomalies are evaluated by an ablation study. Finally, the results are placed in a broader perspective by comparing DR{\AE}M with state-of-the-art weakly-supervised and fully-supervised surface-defect detection methods.}

\subsection{Comparison with unsupervised methods}

\begin{figure*}[!htb]
\centering
  \includegraphics[width=0.88\linewidth]{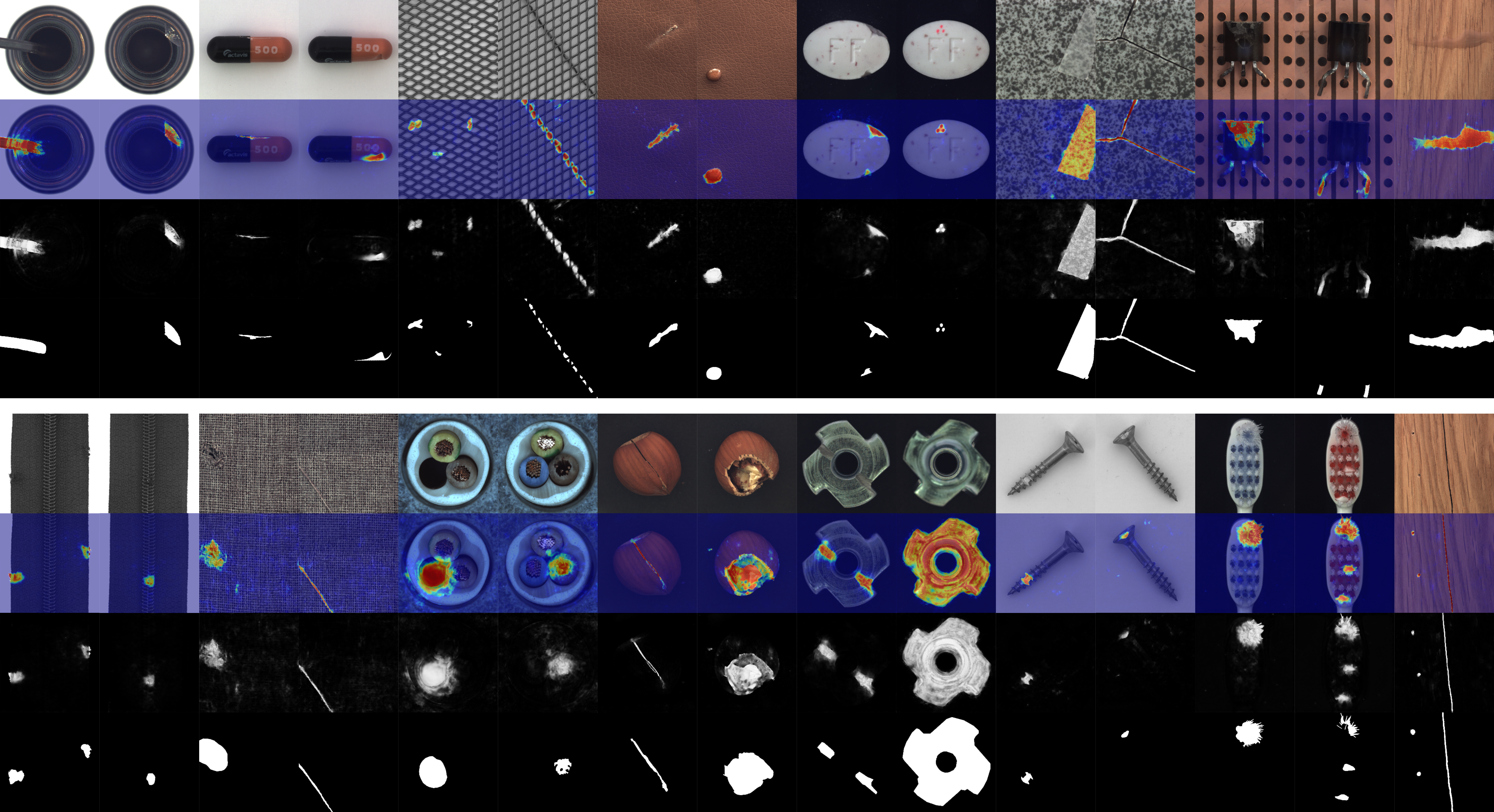}
\caption{Qualitative examples. The original image, the anomaly map overlay, the anomaly map and the ground truth map are shown.}
\label{fig:qualitative}
\end{figure*}

DR{\AE}M is evaluated on the recent challenging MVTec anomaly detection dataset \cite{bergmann2019mvtec}, which has been established as a standard benchmark dataset for evaluating unsupervised surface anomaly detection methods. We evaluate DR{\AE}M on the tasks of surface anomaly detection and localisation. The MVTec dataset contains 15 object classes with a diverse set anomalies which enables a general evaluation of surface anomaly detection methods. Anomalous examples of the MVTec dataset are shown in Figure~\ref{fig:qualitative}. For evaluation, the standard metric in anomaly detection, AUROC, is used. Image-level AUROC is used for anomaly detection and a pixel-based AUROC for evaluating anomaly localization \cite{bergmann2018improving, schlegl2017unsupervised,liu2018future, tang2020anomaly}. 
\ntext{The AUROC, however, does not reflect the localization accuracy well in surface anomaly detection setups, where only a small fraction of pixels are anomalous. The reason is that false positive rate is dominated by the a-priori very high number of non-anomalous pixels and is thus kept low despite of false positive detections.
We thus additionally report the pixel-wise average precision metric (AP), which is more appropriate for highly imbalanced classes and in particular for surface anomaly detection, where the precision plays an important role.}
 

In our experiments, the network is trained for 700 epochs on the MVTec anomaly detection dataset \cite{bergmann2019mvtec}. The learning rate is set to $10^{-4}$ and is multiplied by $0.1$ after $400$ and $600$ epochs. Image rotation in the range of $(-45,45)$ degrees is used as a data augmentation method on anomaly free images during training to alleviate overfitting due to the relatively small anomaly-free training set size. The Describable Textures Dataset \cite{cimpoi2014describing} is used as the anomaly source dataset.

A number of obtained qualitative examples are presented in Figure~\ref{fig:qualitative}. As one can observe, the obtained anomaly masks are very detailed and resemble the given ground truth labels to a high degree of accuracy. Consequently, DR{\AE}M achieves state-of-the-art quantitative results across all MVTec classes for surface anomaly detection as well as localization. 

\begin{figure}
\centering
  \includegraphics[width=0.75\linewidth]{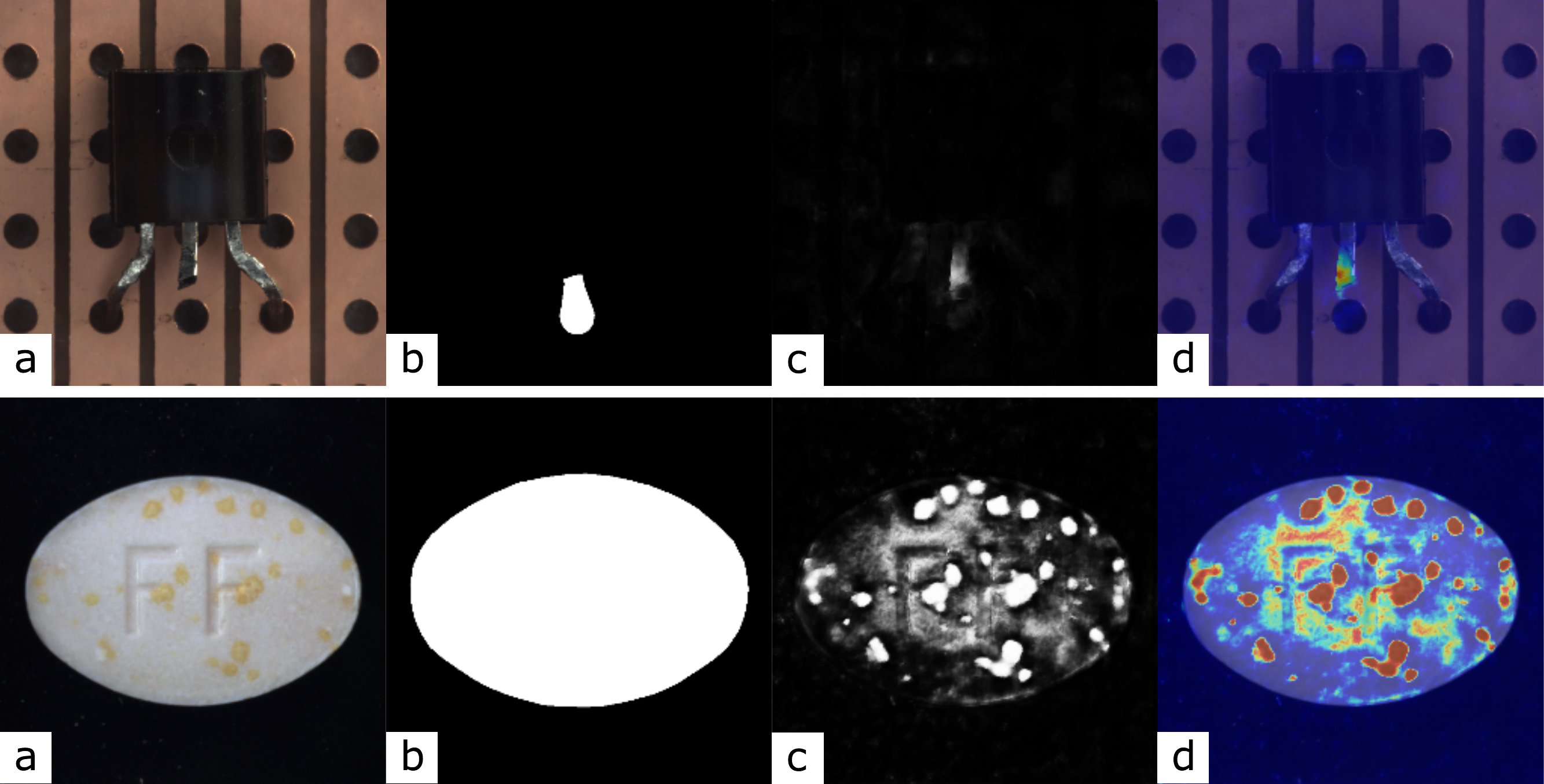}
\caption{The original image (a) contains anomalies which are difficult to mark in the ground truth mask (b) which causes a discrepancy between the ground truth and the output anomaly map (c,d).}
\label{fig:tran_annot}
\end{figure}

\textbf{Surface Anomaly Detection.} 
\ntext{Table \ref{tab:resSota} quantitatively compares DR{\AE}M with recent approaches on the task of \emph{image-level surface anomaly detection}.}
\ntxt{DR{\AE}M significantly outperforms all recent surface anomaly detection methods, achieving the highest AUROC in $9$ out of $15$ classes and achieving comparable results in the other classes. It surpasses the previous best state-of-the-art approach by $2.5$ percentage points. The reduced performance in some classes could be explained by particularly difficult anomalies that are close to the normal image distribution. The absence of a part of the object is especially difficult to detect. Regions, where the object features are missing, usually contain other commonly occurring features. This makes such anomalies difficult to distinguish from anomaly-free regions. An example of this can be seen in Figure \ref{fig:tran_annot}, where some of the transistor leads had been cut. The ground truth marks the area where the broken lead should be as anomalous. DR{\AE}M only detects anomalous features in a small region of the cut lead, as the background features are common during training.}

\begin{table}
\centering
\resizebox{0.95 \linewidth}{!}{\begin{tabular}{c c c c c c c c}
\hline
Class & \cite{akcay2018ganomaly} & \cite{tang2020anomaly} & \cite{bergmann2020uninformed} & \cite{zavrtanik2020riad} & \cite{madGauss} & \cite{defard2020padim} & DR{\AE}M \\ \hline \hline
bottle     & 79.4 & 98.3 & 99.0 & 99.9 & \textbf{100}  & 99.9 & 99.2   \\ 
capsule    & 72.1 & 68.7 & 86.1 & 88.4 & 92.3 & 91.3 & \textbf{98.5}  \\ 
grid       & 74.3 & 86.7 & 81.0 & 99.6 & 92.9 & 96.7 & \textbf{99.9}  \\ 
leather    & 80.8 & 94.4 & 88.2 & \textbf{100}  & \textbf{100}  & \textbf{100}  & \textbf{100}   \\ 
pill       & 67.1 & 76.8 & 87.9 & 83.8 & 83.4 & 93.3 & \textbf{98.9}   \\ 
tile       & 72.0 & 96.1 & 99.1 & 98.7 & 97.4 & 98.1 & \textbf{99.6}  \\ 
transistor & 80.8 & 79.4 & 81.8 & 90.9 & 95.9 & \textbf{97.4} & 93.1  \\ 
zipper     & 74.4 & 78.1 & 91.9 & 98.1 & 97.9 & 90.3 & \textbf{100}   \\ 
cable      & 71.1 & 66.5 & 86.2 & 81.9 & \textbf{94.0} & 92.7 & 91.8  \\ 
carpet     & 82.1 & 90.3 & 91.6 & 84.2 & 95.5 & \textbf{99.8} & 97.0  \\ 
hazelnut   & 87.4 & 100  & 93.1 & 83.3 & 98.7 & 92.0 & \textbf{100.0}  \\ 
metal nut  & 69.4 & 81.5 & 82.0 & 88.5 & 93.1 & \textbf{98.7} & \textbf{98.7}  \\ 
screw      & \textbf{100}  & \textbf{100}  & 54.9 & 84.5 & 81.2 & 85.8 & 93.9  \\ 
toothbrush & 70.0 & 95.0 & 95.3 & \textbf{100}  & 95.8 & 96.1 & \textbf{100}   \\ 
wood       & 92.0 & 97.9 & 97.7 & 93.0 & 97.6 & \textbf{99.2} & 99.1 \\ \hline 
$avg$      & 78.2 & 87.3 & 87.7 & 91.7 & 94.4 & 95.5 & \textbf{98.0}  \\ \hline \\
\end{tabular}}
\caption{Results for the task of surface anomaly detection on the MVTec dataset (AUROC). An average score over all classes is also reported the last row ($avg$).}
\label{tab:resSota}
\end{table}

\begin{figure}
\centering
  \includegraphics[width=0.70\linewidth]{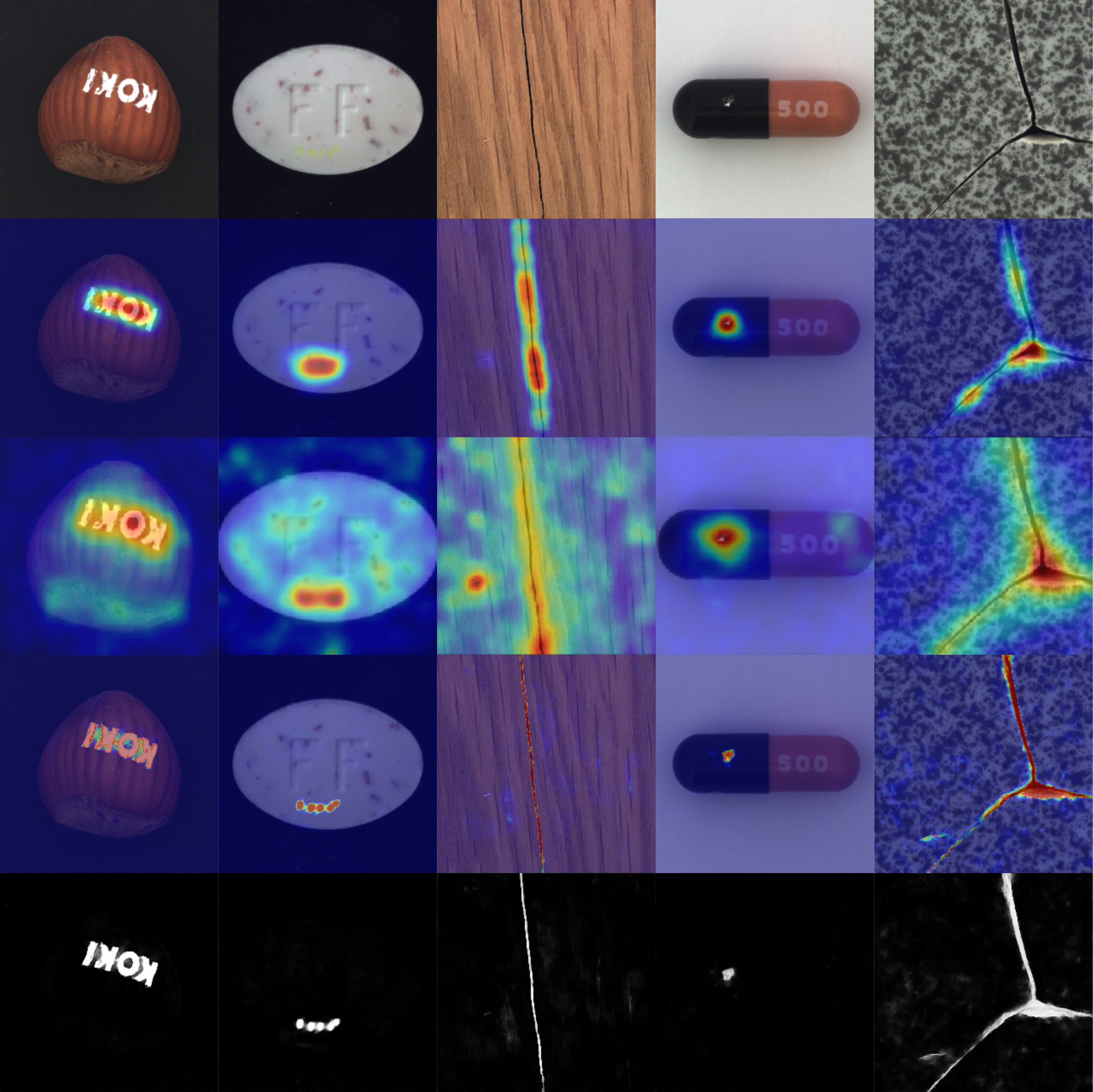}
\caption{The anomalous images are shown in the first row. The middle three rows show the anomaly maps generated by our implementation of Uninformed Students~\cite{bergmann2020uninformed}, PaDim~\cite{defard2020padim} and DR{\AE}M, respectively. The last row shows the direct anomaly map output of DR{\AE}M.}
\label{fig:us_comp}
\end{figure}

\textbf{Anomaly Localization.} \ntext{Table \ref{tab:resSotaLoc} compares DR{\AE}M to the recent state-of-the-art  on the task of \emph{pixel-level surface anomaly detection}.} 
\ntxt{DR{\AE}M achieves 
comparable results to the previous best-performing methods in terms of AUROC scores and surpasses the state-of-the-art by $13.4$ percentage points in terms of AP. A better AP score is achieved in $11$ out of $15$ classes and is comparable to the state-of-the-art in other classes. A qualitative comparison with the state-of-the-art method Uninformed Students~\cite{bergmann2020uninformed} and PaDim~\cite{defard2020padim} is shown in Figure \ref{fig:us_comp}. DR{\AE}M achieves a significant improvement in anomaly segmentation accuracy. }

\ntext{A detailed inspection showed that some of the detection errors can be attributed to the inaccurate ground truth labels on ambiguous anomalies.} An example of this is shown in Figure \ref{fig:tran_annot}, where the ground truth covers the entire surface of the pill, yet only the yellow dots are anomalous. DR{\AE}M produces an anomaly map that correctly localizes the yellow dots, but the discrepancy with the ground truth mask increases the performance error. \ntext{These annotation ambiguities also impact the AP score of the evaluated methods.}


\begin{table}[h]
\centering
\resizebox{0.95\linewidth}{!}{\begin{tabular}{c c c c c}
\hline
Class  &  US\cite{bergmann2020uninformed} &  RIAD\cite{zavrtanik2020riad} & PaDim\cite{defard2020padim} & DR{\AE}M \\ \hline \hline
bottle        & 97.8 / 74.2 &  98.4 / 76.4 & 98.2 / 77.3 & \textbf{99.1} / \textbf{86.5} \\ 
capsule       & 96.8 / 25.9 &  92.8 / 38.2 & \textbf{98.6} / 46.7 & 94.3 / \textbf{49.4} \\ 
grid          & 89.9 / 10.1 &  98.8 / 36.4 & 97.1 / 35.7 & \textbf{99.7} / \textbf{65.7} \\ 
leather       & 97.8 / 40.9 &  99.4 / 49.1 & \textbf{99.0} / 53.5 & 98.6 / \textbf{75.3} \\ 
pill          & 96.5 / \textbf{62.0} &  95.7 / 51.6 & 95.7 / 61.2 & \textbf{97.6} / 48.5 \\ 
tile          & 92.5 / 65.3 &  89.1 / 52.6 & 94.1 / 52.4 & \textbf{99.2} / \textbf{92.3} \\ 
transistor    & 73.7 / 27.1 &  87.7 / 39.2 & \textbf{97.6} / \textbf{72.0} & 90.9 / 50.7 \\ 
zipper        & 95.6 / 36.1 &  97.8 / 63.4 & 98.4 / 58.2 & \textbf{98.8} / \textbf{81.5} \\ 
cable         & 91.9 / 48.2 &  84.2 / 24.4 & \textbf{96.7} / 45.4 & 94.7 / \textbf{52.4} \\ 
carpet        & 93.5 / 52.2 &  96.3 / \textbf{61.4} & \textbf{99.0} / 60.7 & 95.5 / 53.5 \\ 
hazelnut      & 98.2 / 57.8 &  96.1 / 33.8 & 98.1 / 61.1 & \textbf{99.7} / \textbf{92.9} \\ 
metal nut     & 97.2 / 83.5 &  92.5 / 64.3 & 97.3 / 77.4 & \textbf{99.5} / \textbf{96.3} \\ 
screw         & 97.4 / 7.8  &  \textbf{98.8} / 43.9 & 98.4 / 21.7 & 97.6 / \textbf{58.2} \\ 
toothbrush    & 97.9 / 37.7 &  \textbf{98.9} / 50.6 & 98.8 / \textbf{54.7} & 98.1 / 44.7 \\ 
wood          & 92.1 / 53.3 &  85.8 / 38.2 & 94.1 / 46.3 & \textbf{96.4} / \textbf{77.7} \\ \hline 
$avg$         & 93.9 / 45.5 &  94.2 / 48.2 & \textbf{97.4} / 55.0 & 97.3 / \textbf{68.4} \\ \hline \\
\end{tabular}}
\caption{Results for the task of anomaly localization on the MVTec dataset (AUROC / AP).}
\label{tab:resSotaLoc}
\end{table}

\subsection{Ablation Study}

The DR{\AE}M design choices are analyzed by groups of experiments evaluating (i) the method architecture, \ntxt{(ii) the choice of anomaly appearance patterns and (iii) low perturbation example generation.} Results are visually grouped by shades of gray in Table \ref{tab:various_ablations}.

\begin{table*}
\centering
\resizebox{0.88\linewidth}{!}{\begin{tabular}{c c c c c  c c  c c c c}
\hline
 & \multicolumn{2}{c}{Architecture}   & \multicolumn{6}{c}{Anomaly Generation}  & \multicolumn{2}{c}{Results} \\  
Method                & Recon. Net. & Discr. Net. & Augmentation & $\beta$    & ImageNet   & DTD        & Perlin     & Rectangle         & Det. & Loc.  \\ \hline 
Disc.                 &             & \checkmark         &  \checkmark  & \checkmark &            & \checkmark & \checkmark &            &  93.9 & 92.7 / 62.5 \\ 
Recon.-AE             & \checkmark  &                   &  \checkmark  & \checkmark &            & \checkmark & \checkmark &             &  83.9 & 89.7 / 47.5     \\ 
Recon.-AE$_{MSGMS}$   & \checkmark  &                    &  \checkmark  & \checkmark &            & \checkmark & \checkmark &            &  90.7 & 93.4 / 50.9      \\ 
Bo\v{z}i\v{c} \etal~\cite{bozic2020end}     &             &   &  \checkmark  & \checkmark &            & \checkmark & \checkmark &            &  92.8 & 93.9 / 60.7 \\ 
\rowcolor{Gray}
DR{\AE}M$_{ImageNet}$ & \checkmark  &     \checkmark     &  \checkmark  & \checkmark & \checkmark &            & \checkmark &            & 97.9  & 97.0 / 67.9   \\ 
\rowcolor{Gray}
DR{\AE}M$_{color}$              & \checkmark  &     \checkmark     &    & \checkmark &            &  & \checkmark &            & 96.2 & 92.6 / 56.5   \\ \rowcolor{Gray}
DR{\AE}M$_{rect}$     & \checkmark  &     \checkmark     &  \checkmark  & \checkmark &            & \checkmark &            & \checkmark & 96.9 & 96.8 / 65.1   \\ 

\rowcolor{Gray2}
DR{\AE}M$_{no\_aug}$  & \checkmark  &     \checkmark     &              &            &            & \checkmark & \checkmark &            &  97.4 & 94.5 / 64.3  \\ 
\rowcolor{Gray2}
DR{\AE}M$_{img\_aug}$ & \checkmark  &     \checkmark     &  \checkmark  &            &            & \checkmark & \checkmark &            &  97.4 & 95.0 / 64.5  \\ 
\rowcolor{Gray2}
DR{\AE}M$_{\beta}$    & \checkmark  &     \checkmark     &              & \checkmark &            & \checkmark & \checkmark &            & 97.9 & 97.1 / \textbf{68.4}   \\ 

\rowcolor{Gray3}
DR{\AE}M              & \checkmark  &     \checkmark     &  \checkmark  & \checkmark &            & \checkmark & \checkmark &            & \textbf{98.0} & \textbf{97.3} / \textbf{68.4}   \\ \hline \\

\end{tabular}}
\caption{Surface anomaly detection (Det.) and localization (Loc.) experiments of the ablation study grouped by shades of gray into (i) method architecture, (ii) anomaly source dataset, (iii) hard simulated anomaly generation, (iv) simulated anomaly shape, and (v) the performance of DR{\AE}M for reference.}
\label{tab:various_ablations}
\end{table*}

\textbf{Architecture.} 
The DR{\AE}M reconstructive sub-network impact on the downstream surface anomaly detection performance is evaluated by removing it from the pipeline and training the discriminative sub-network alone. The results are shown in Table~\ref{tab:various_ablations}, experiment Disc. Note a reduction in performance in comparison to the full DR{\AE}M architecture (Table \ref{tab:various_ablations}, experiment DR{\AE}M). \ntext{The performance drop is due to overfitting of the discriminative sub-network to the simulated anomalies, which are not a faithful representation of the real ones.}


Next, the discriminative power of the reconstructive sub-network alone is analyzed by evaluating it as an auto-encoder-based surface anomaly detector. The reconstructed image output of the sub-network is compared to the input image using the SSIM function~\cite{wang2004image} to generate the anomaly map. The results of this approach are shown in Table \ref{tab:various_ablations}, experiment Recon.-AE. Recon.-AE outperforms the recent auto-encoder-based surface anomaly detection method AE-SSIM\cite{bergmann2018improving} (see results in Table~\ref{tab:resSotaLoc})
This suggests that simulated anomaly training introduces additional information into the auto-encoder-based training, but judging by the performance gap to DR{\AE}M, the SSIM similarity function may not be optimal for extraction of the anomaly information. 
Indeed, using the recently proposed similarity function MSGMS~\cite{zavrtanik2020riad} (Recon.-AE$_{MSGMS}$) improves the performance, but the results are still significantly worse than when using the entire DR{\AE}M architecture, which indicates that both reconstructive and discriminative parts are required for optimal results. 

To further emphasize the contribution of the DR{\AE}M backbone, we replace it entirely by the recent state-of-the-art supervised discriminative surface anomaly detection network~\cite{bozic2020end} and re-train with the simulated anomalies (Table \ref{tab:various_ablations}, Bo\v{z}i\v{c} \etal). \ntext{Performance substantially drops, which further supports the power of learning the \textit{anomaly deviation extent} from normality rather than the anomaly or normality \textit{appearance}.}

\begin{figure}
\centering
  \includegraphics[width=0.85\linewidth]{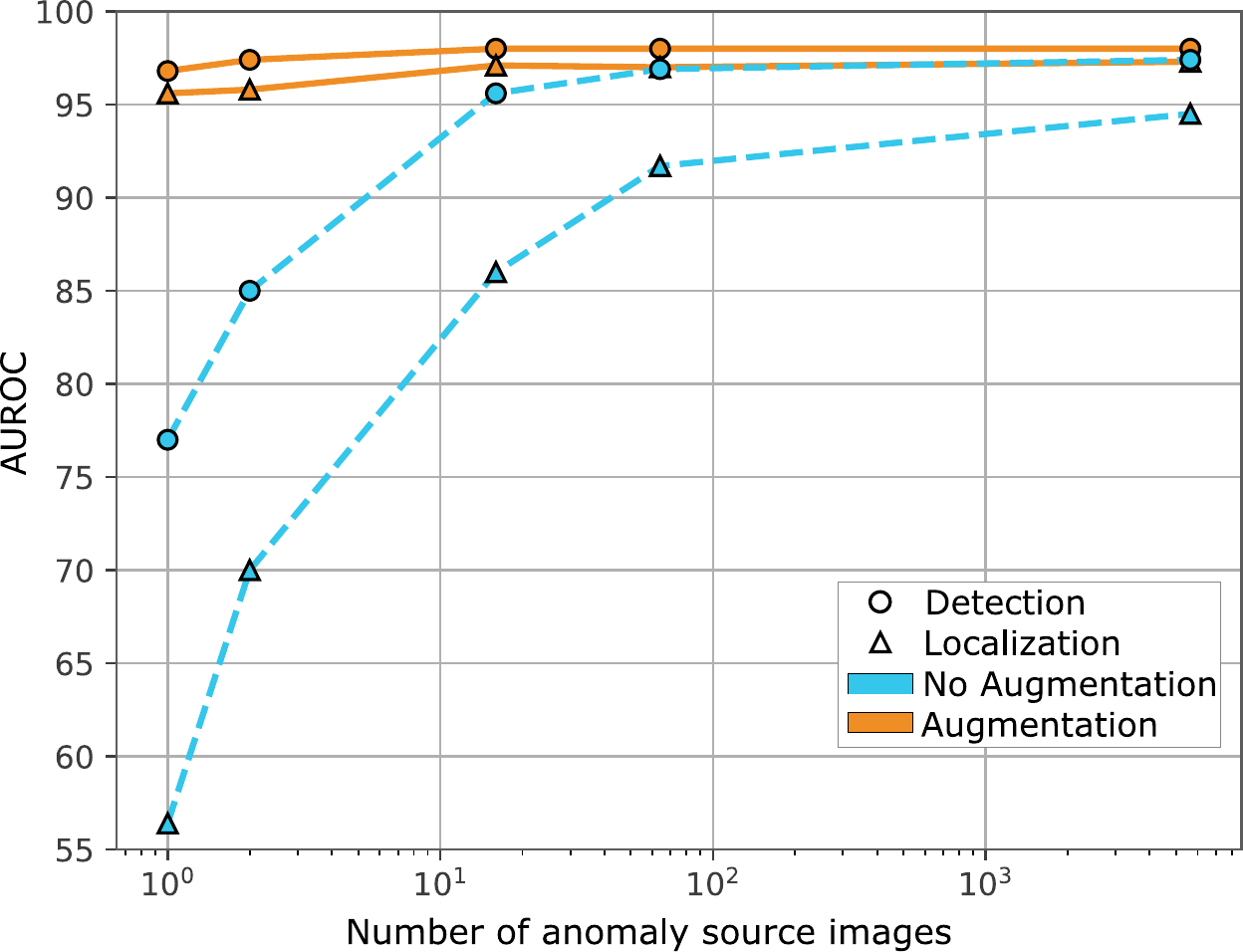}
\caption{DR{\AE}M achieves a remarkable detection and localization performance already at as low as 10 texture source images in the simulator when augmentation is applied.} 
\label{fig:src_num}
\end{figure}

\textbf{Anomaly appearance patterns.} 
\ntext{DR{\AE}M is re-trained using ImageNet~\cite{deng2009imagenet} as the texture source in the anomaly simulator to study the influence of the anomaly generation dataset (DR{\AE}M$_{ImageNet}$ in Table~\ref{tab:various_ablations}).} 
Results are comparable to using the much smaller DTD~\cite{cimpoi2014describing} dataset. \ntxt{Figure~\ref{fig:src_num} shows the performance at various anomaly source dataset sizes. Results suggest that the augmentation and opacity randomization substantially contribute to performance allowing remarkably small number of texture images (less than 10).} \ntext{As an extreme case, the anomaly textures are generated as homogeneous regions of a randomly sampled color (DR{\AE}M$_{color}$). Note that DR{\AE}M$_{color}$ still achieves state-of-the-art results, further suggesting that DR{\AE}M does not require simulations to closely match the real anomalies.}

\begin{figure}
\centering
  \includegraphics[width=0.8\linewidth]{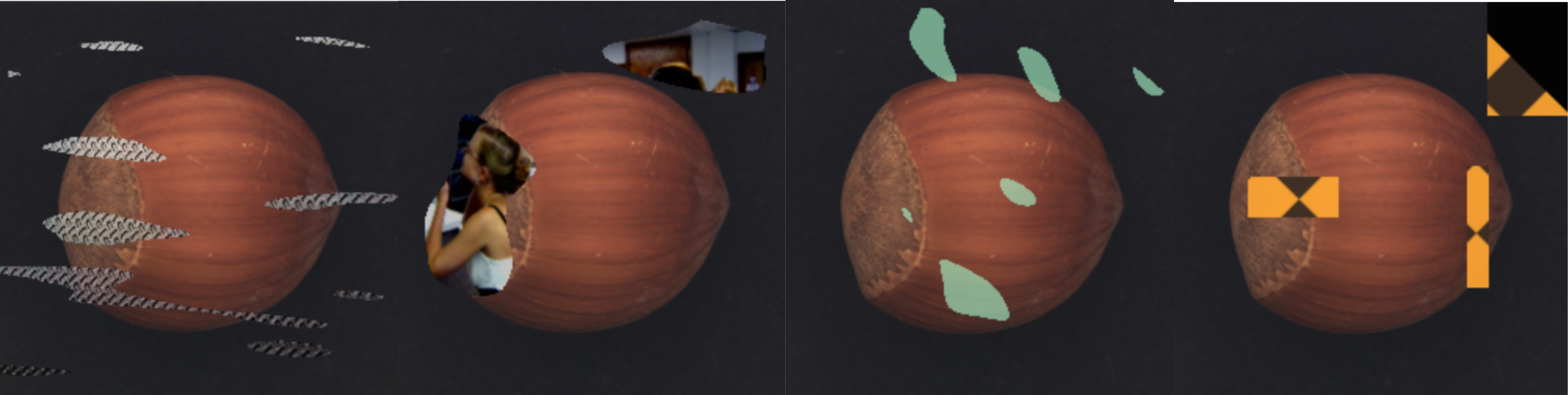}
\caption{Anomalies simulated using the DTD~\cite{cimpoi2014describing} (DR{\AE}M), ImageNet~\cite{deng2009imagenet} (DR{\AE}M$_{ImageNet}$), homogeneous color regions (DR{\AE}M$_{color}$) and rectangular masks (DR{\AE}M$_{rect}$), from left to right.}
\label{fig:anom_app}
\end{figure}

\ntext{The impact of the anomaly shape generator is evaluated by replacing the Perlin noise generator by a rectangular region generator.}
The anomaly mask is thus generated by sampling multiple rectangular areas for the anomalous regions (DR{\AE}M$_{rect}$ in Table \ref{tab:various_ablations}). Training on rectangular anomalies causes only a slight performance drop and suggests that the simulated anomaly shape does not have to be realistic to generalize well to real world anomalies. 
\ntext{Examples of anomalies generated in anomaly appearance ablation experiments are shown in Figure~\ref{fig:anom_app}.}

\textbf{Low perturbation examples.} \ntext{The anomaly source image augmentation and the opacity randomization are responsible for tightening the decision boundary around the anomaly-free training distribution. Table~\ref{tab:various_ablations} reports the results of DR{\AE}M variants trained (i) 
without image augmentation and opacity randomization (DR{\AE}M$_{no\_aug}$), (ii) using only image augmentation (DR{\AE}M$_{img\_aug}$) and (iii) using only opacity randomization (DR{\AE}M$_{\beta}$).}
There is a significant localization performance gap between DR{\AE}M$_{no\_aug}$ and DR{\AE}M, however, this can be significantly narrowed by using the opacity randomization in training even without image data augmentation. 

%

\subsection{Comparison with supervised methods}

Supervised methods require anomaly annotations at training time and cannot be evaluated on MVTec. We thus compare DR{\AE}M with the supervised methods on the DAGM dataset~\cite{dagm2007}
that contains 10 textured object classes with small anomalies visually very similar to the background, which makes the dataset particularly challenging for the unsupervised methods.

 
DR{\AE}M is trained only on anomaly-free training samples using the same parameters as in previous experiments.
The standard evaluation protocol on this dataset~\cite{racki2018, zhang2020, lin2020efficient, bozic2020end} is used -- the challenge is to classify whether the image contains the anomaly; localization accuracy is not measured, since the anomalies are only coarsely labeled.
 



Table \ref{tab:resDAGM} shows that the best fully supervised methods 
nearly perfectly classify anomalous images,
while the state-of-the-art unsupervised methods like RIAD~\cite{zavrtanik2020riad} and US~\cite{bergmann2020uninformed} 
struggle with subtle anomalies on highly textured regions\footnote{Please see the supplementary material for additional qualitative results.}.
DR{\AE}M significantly outperforms these methods, and even the weakly supervised CADN~\cite{zhang2020} by a large margin, obtaining classification performance close to the best fully-supervised methods, which is a remarkable result. 

\begin{figure}
\centering
  \includegraphics[width=0.85\linewidth]{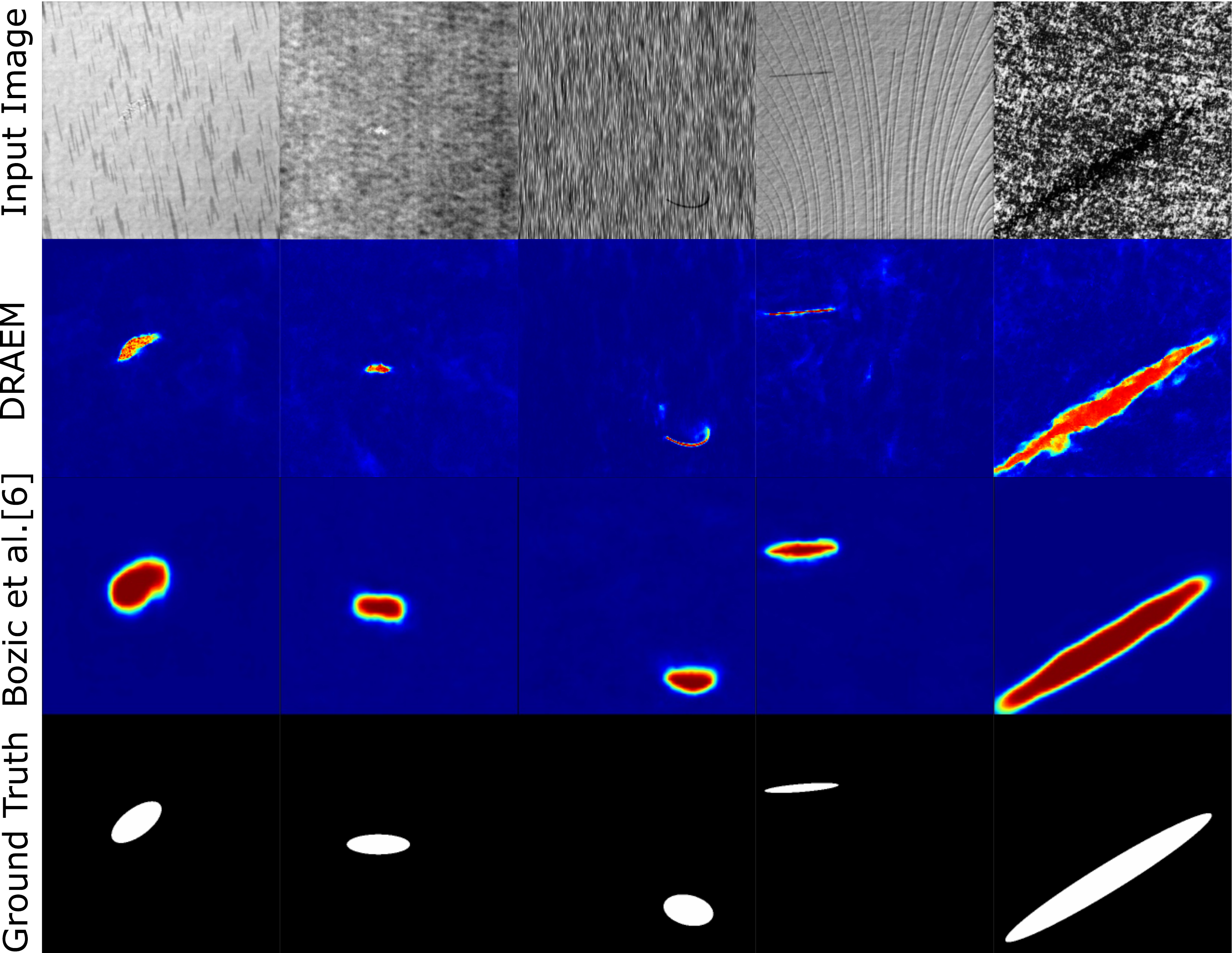}
\caption{Supervised methods replicate the approximate ground truth training annotations, leading to a low localization accuracy. DR{\AE}M does not use the ground truth, yet produces far better localization.}
\label{fig:dagm_qual}
\end{figure}

Furthermore, DR{\AE}M outperforms all supervised methods in terms of anomaly localization accuracy on this dataset. Since the training images are only coarsely annotated with ellipses that approximately cover the surface defects and contain background, the supervised methods produce inaccurate localization in test images as well.
In contrast, DR{\AE}M does not use the labels at all, and thus 
produces more accurate anomaly maps, as shown in Figure~\ref{fig:dagm_qual}.



\begin{table}
\centering
\resizebox{0.75\linewidth}{!}{\begin{tabular}{c  c c c c c}
\hline
  &   Methods & AUROC & TPR  & TNR  & CA   \\  \hline
\multirow{3}{*}{\sidetext{\rotatebox[origin=c]{90}{Unsup.}}} & RIAD~\cite{zavrtanik2020riad}      &  78.6  & 79.2 & 69.1 & 70.4  \\     
& US~\cite{bergmann2020uninformed}        &  72.5  & 72.6 & 65.3 & 66.2 \\
& MAD~\cite{madGauss}        &  82.4  & 78.7 & 85.7 & 66.2 \\
& PaDim~\cite{defard2020padim}        &  95.0  & 83.3 & 97.5 & 95.7 \\
& DR{\AE}M  &  \textbf{99.0}  & \textbf{96.5} & \textbf{99.4} & \textbf{98.5}  \\ \hline
\multirow{4}{*}{\sidetext{\rotatebox[origin=c]{90}{Sup.}}} & CADN~\cite{zhang2020}     &   -    &  -   &   -  & 89.1  \\ 
& Ra\v{c}ki \etal~\cite{racki2018}     &  99.6  & 99.9 & 99.5 &  -  \\
& Lin \etal~\cite{lin2020efficient}     &  99.0  & 99.4 & 99.9 &  -    \\
& Bo\v{z}i\v{c} \etal~\cite{bozic2020end}     &   \textbf{100}    & \textbf{100}  & \textbf{100}  &  \textbf{100}    \\ \hline
\end{tabular}}
\caption{
DR{\AE}M outperforms unsupervised methods on DAGM dataset and performs on par with fully supervised ones.
}
\label{tab:resDAGM}
\end{table}

\section{Conclusion}
A discriminative end-to-end trainable \ntext{surface} anomaly detection and localization method DR{\AE}M was presented. 
\ntxt{DR{\AE}M outperforms the current state-of-the-art on the MVTec dataset~\cite{bergmann2019mvtec} by $2.5$ AUROC points on the surface anomaly detection task and by $13.5$ AP points on the localization task.} On the DAGM dataset~\cite{dagm2007}, DR{\AE}M delivers anomalous image classification accuracy close to fully supervised methods, while outperforming them in localization accuracy. This is a remarkable result since DR{\AE}M is not trained on real anomalies. In fact, a detailed analysis shows that our paradigm of learning a joint reconstruction-anomaly embedding through a reconstructive sub-network significantly improves the results over standard methods and that an accurate decision boundary can be well estimated by \ntext{learning the extent of deviation from reconstruction on simple simulations rather than learning either the normality or real anomaly appearance.} \\



{\small
\bibliographystyle{ieee_fullname}
\bibliography{iccv}
}

\end{document}